\documentclass{article}

\usepackage{csagh}
\usepackage{graphicx}
\usepackage{subfiles}
\usepackage{algorithmicx}
\usepackage{algorithm}
\usepackage{algpseudocode}
\usepackage{esvect}
\usepackage{mathrsfs}
\usepackage{comment}
\usepackage[none]{hyphenat}
\usepackage{amsmath}
\usepackage{amssymb}
\usepackage{bm}
\usepackage[combine]{ucs}
\usepackage[utf8x]{inputenc}
\begin{document}
\begin{opening}

\title{A Survey on Multi-Objective Based Parameter Optimization for Deep Learning}
\author[Jadavpur University, Kolkata,India, mrittikac.cse.rs@jadavpuruniversity.in]{Mrittika Chakraborty}
\author[Jadavpur University, Kolkata,India,wreetbhaspal@gmail.com]{Wreetbhas Pal}
\author[Indian Statistical Institute, Kolkata,India, sanghami@isical.ac.in]{Sanghamitra Bandyopadhyay, FIEEE}
\author[Jadavpur University, Kolkata,India,ujjwal.maulik@jadavpuruniversity.in,* -Corresponding Author]{Ujjwal Maulik*, FIEEE}

\begin{abstract}
Deep learning models form one of the most powerful machine learning models for the extraction of important features. Most of the designs of deep neural models, i.e., the initialization of parameters, are still manually tuned. Hence, obtaining a model with high performance is exceedingly time-consuming and occasionally impossible. Optimizing the parameters of the deep networks, therefore, requires improved optimization algorithms with high convergence rates. The single objective-based optimization methods generally used are mostly time-consuming and do not guarantee optimum performance in all cases. Mathematical optimization problems containing multiple objective functions that must be optimized simultaneously fall under the category of multi-objective optimization sometimes referred to as Pareto optimization. Multi-objective optimization problems form one of the alternatives yet useful options for parameter optimization. However, this domain is a bit less explored. In this survey, we focus on exploring the effectiveness of multi-objective optimization strategies for parameter optimization in conjunction with deep neural networks. The case studies used in this study focus on how the two methods are combined to provide valuable insights into the generation of predictions and analysis in multiple applications.
\end{abstract}

\keywords{deep learning; multi-objective optimization; parameter optimization; neural networks}

\end{opening}

\section{Introduction}
Deep learning \cite{r9} is a particular kind of machine learning using artificial neural networks and motivated by the idea of information processing in biological systems. Deep learning helps in classification, detection, segmentation, etc. Deep neural networks (DNNs) are made up of several layers of interconnected nodes, and they learn to perform tasks through the adjustment of the network's parameters.\\
Hyperparameters of DNNs include the number of layers, the number of neurons in each layer, the activation functions, the learning rate, the optimizer, the loss function, and the batch size, among others. These hyperparameters need to be set up manually. Therefore, developing a deep learning model for various types of problems is a complex task that requires significant effort, as numerous parameters require to be fine-tuned. Typically, these models are developed using the knowledge of skilled experts. However, over the past few years, there is a surge in research for designing deep learning architectures using optimization techniques. Initially, optimization methods such as grid search \cite{r75}, random search \cite{r10}, and Bayesian optimization \cite{r11} were used for parameter optimization. Along with these techniques, single-objective optimization algorithms have also been employed to achieve high levels of classification accuracy \cite{r12, r13,r14,r15,r16,r17,r18,r19}. Since optimization of more than one objective is needed, multi-objective optimization methods were introduced. For instance, the objective could be to obtain the maximum accuracy with the least parameters or to do so in the least amount of time \cite{r20,r21,r22,r23}.\\
The method of optimization involves determining the optimal or best solution, which can be done by searching for maximum or minimum values, using single or multiple objectives. When a problem has more than one objective, it is known as Multi-Objective Optimization (MOO). MOO has many applications in the real world, including mechanics, politics, finance, and economics. For example, in the field of mechanics \cite{r7,r8}, MOO can be used to minimize the total cost of tube heat exchangers and shells, including annual energy expenditure and capital investment while reducing the heat exchanger's length using a genetic algorithm (GA) \cite{r78}. MOO algorithms are designed to find optimal values for variables such as baffle spacing, outer diameter, and outer tube diameter. In politics \cite{r6}, MOO can be used to figure out important players who gain from political campaigns, while in finance \cite{r2,r3,r4,r5}, it can be used to spot noteworthy technical analysis trends in time series of financial data. Advancements in the field of biotechnology have also been cited in \cite{r1}. Here, MOO has been applied to optimize the fisheries bio-economic model by minimizing waste, maintaining quota shares, and maximizing profits.\\
There are different settlement methods used for solving MOO problems. One of these methods is the global criterion method \cite{r45}, which aims to transform multiple optimization problems into a single optimization problem by reducing the gap between several reference points and feasible solutions. Another method is the weighted-sum method \cite{r47,r48,r49,r50,r51,r52,r53}, which combines all the problems into a single problem using a weighted vector. However, choosing the weights for problems with different magnitudes can be challenging and lead to bias. In cases where the plural problem being optimized is not convex, the $\epsilon$-constraint method \cite{r54} is utilized. This approach optimizes one problem while transforming other problems into constraints or restrictions.\\
The lexicographic method \cite{r55} is used to optimize objectives by prioritizing their order of importance. Each objective is optimized individually, starting with the most vital goal. If there is only one solution returned, it is considered the best solution. If not, optimization continues on the next objective under new restrictions based on the solution from the first objective. The goal programming method \cite{r56,r57,r59,r60,r61,r62} involves determining the objective function's ambition level to be achieved. Multi-objective evolutionary algorithms (MOEAs) \cite{r77,r63} are stochastic optimization techniques used to find optimal Pareto solutions. Most of the time, MOEAs use dominance relationships in their actions, and their optimization mechanism is similar to that of evolutionary algorithms. MOEAs can also apply conventional support techniques such as niching due to the existence of objective space. Various MOO settlement methods have been reviewed, which involve solving complex equations.\\
In this paper, survey work is done on multi-objective based parameter optimization for deep learning focused on various applications like healthcare, language processing, machinery, and others.
\section{Deep Learning}
\begin{figure}[h]
\centering
\includegraphics[scale=0.3]{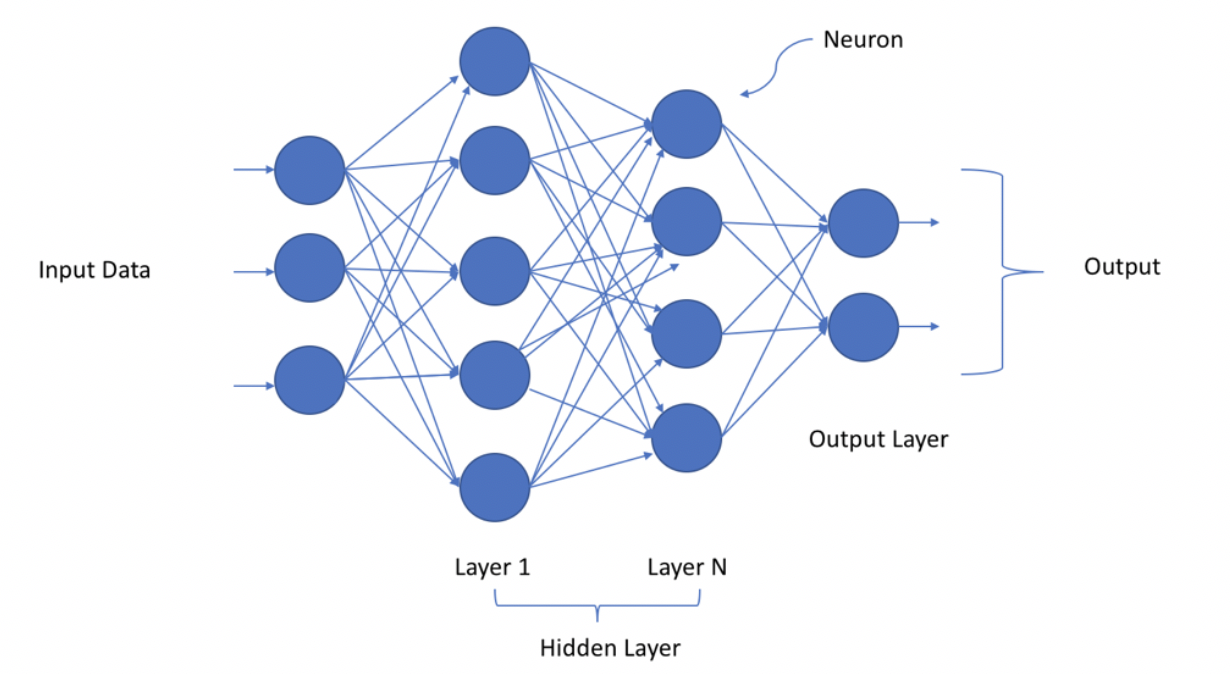}
\caption{\centering A Deep Neural Network with `N' hidden layers}\label{fig:fig5}
\end{figure}

Deep learning is a particular kind of machine learning that utilizes artificial neural networks (ANNs) \cite{r113, r114, r103} with multiple layers, also known as deep neural networks (DNNs) \cite{r110, r111, r112, r104}. These networks are designed to model and process non-linear relationships, taking inspiration from the anatomy and physiology of the human brain. They are capable of learning from vast amounts of data in an unsupervised or semi-supervised manner. The layering of neurons makes up a DNN in its compacted form, where the neurons resemble those found in the brain and connections between them. These neurons receive input, process it, and pass on signals to other neurons, forming a sophisticated network that improves with experience. The given diagram Figure.~\ref{fig:fig5} illustrates a deep neural network (DNN), made up of several layers (`N' layered) of artificial neurons. As it can be seen in Figure.~\ref{fig:fig5}, the first layer's neurons process the input data, which is then passed on to the following layer, and so on until the final output is produced. One or more neurons may be present in each layer, and these neurons compute an extremely tiny function called the activation function. If the incoming neurons' result exceeds a certain cutoff, the output is forwarded to the next connected neuron. There is a weight associated with the connection between two neurons in successive layers that shows the effect of the input on the output. These weights are iteratively changed during model training to discover the best way to predict the desired outcome. The logical building blocks of a neural network include a neuron, layer, weight, input, output, activation function, and a learning mechanism (optimizer) that helps the neural network gradually change the weights for better prediction of the outcome. Deep learning employs several architectures, including feedforward neural networks (FNNs) \cite{r105}, convolutional neural networks (CNNs) \cite{r106}, and recurrent neural networks (RNNs) \cite{r107}. FNNs follow a simple linear information flow through the network and have been employed in natural language processing, spoken word identification, and picture classification. CNNs are specialized FNNs designed specifically for image and video recognition, capable of automatically learning image features and useful for object identifying, picture classifying, and image segmentation tasks \cite{r80}. RNNs are ideal for processing sequential data, like time series and natural language, and can maintain an internal state that captures information about previous inputs, making them well-suited for tasks such as speech recognition, natural language processing, and language translation.
\section{Multi-Objective Optimization(MOO) Methods}
Single-objective (SO) optimization as in \cite{r115, r116, r76} refers to the method of searching for the optimal solution that maximizes or minimizes a single objective function. Although this kind of optimization is a helpful tool for giving decision-makers an understanding of the issue, it frequently does not offer alternate solutions that balance various goals. Instead, it lumps all objectives into one.\\
The Multi-objective optimization (MOO) as suggested by Guantara et.al in \cite{r70} is a process of finding optimal solutions for more than one desired goal. MOO is a useful technique in optimization because it simplifies the problem by not requiring complicated equations. In MOO, decision-making involves a balancing act between various opposing issues. The concept of MOO was created by Vilfredo Pareto in 1896 \cite{r102}, and it involves the objective function vector, where each vector is the solution vector function. There is not a single optimum solution in MOO for all objectives, but they need to be evaluated for trade-offs. \\
The  MOO problem mathematical equation can be stated as follows, \cite{r24}:

\begin{equation} \label{eq1}
\begin{split}
\bm{min/max\ f_1(x),\ f_2(x),....,f_n(x)} \\
\bm{subject\ to\ :\ x\ \epsilon\ \cup;}
\end{split}
\end{equation}

where `$x$', `$n$', $\cup$, $f_n(x)$, and `$min/max$' denotes a solution, the number of objective functions, feasible set, $n^{th}$ objective function, and combined object operations respectively.\\

\begin{figure}[h]
\centering
\includegraphics[scale=0.5]{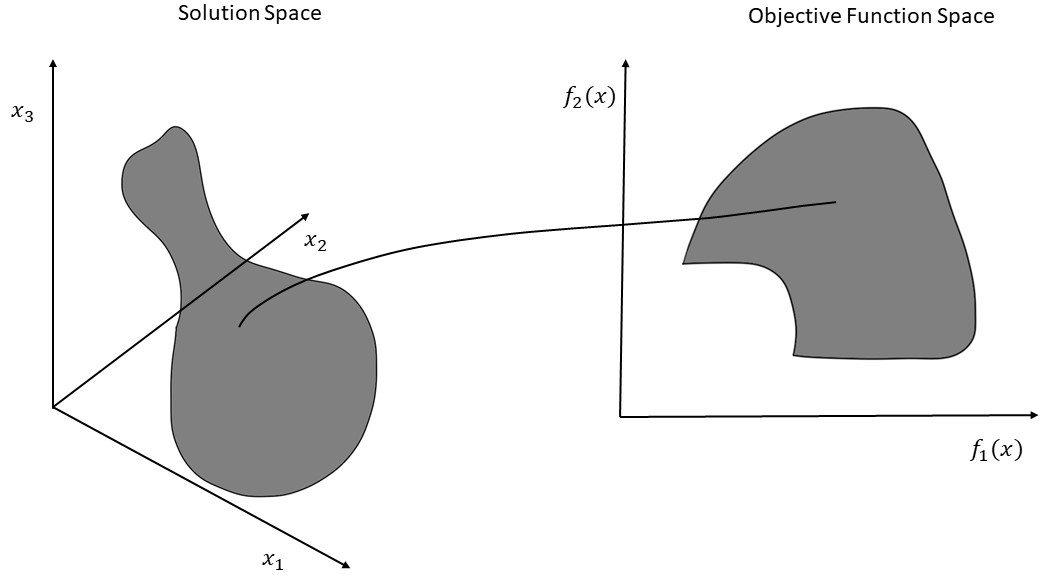}
\caption{\centering Mapping between spaces of solution and objective function \cite{r70}}\label{fig:fig1}
\end{figure}

\begin{figure}[h]
\centering
\includegraphics[scale=0.5]{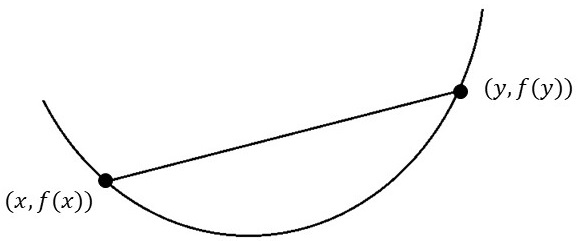}
\caption{\centering Convex function \cite{r70}}\label{fig:fig2}
\end{figure}

A decision variable solution vector and a multi-dimensional objective function vector both have spaces in MOO. There is a corresponding point in the space of the objective function for each solution in the space of the decision variable. The relationship between the two spaces is depicted in Figure.~\ref{fig:fig1} as in \cite{r25}.\\ 
The convexity of the spaces of solution and objective function are important in choosing which algorithm will be employed to solve the problem. If all objective functions and solution regions are convex, MOO problems are considered convex. If the objective function satisfies the equation~\ref{eq2}, then it is convex \cite{r26}:

\begin{equation} \label{eq2}
\bm{f(\theta x+(1-\theta)y)\leq \theta f(x)+(1-\theta)f(y),}
\end{equation}

with the value 1 and f in the $x$, $y$ domain. For a better understanding of equation~\ref{eq2}, it can be deduced that the f graph lies above the line joining $(x,f(x))$ and $(y,f(y))$ from $x$ to $y$. as shown in Figure.~\ref{fig:fig2}. There are two different categories of MOO problem solutions, namely the Pareto method and the Scalarization method \cite{r27}. The Pareto method is employed when the targeted outcomes and performance measures are distinct and get balanced, which is presented in Pareto optimal front (POF) form. On the other hand, the Scalarization method involves using the performance measures to form a scalar function which incorporates the fitness function \cite{r28}. Both methods are explained in the following section.

\subsection{Pareto Method}
The following is a Pareto-based mathematical equation for the MOO problem \cite{r29}:

\begin{equation} \label{eq3}
\begin{split}
\bm{f_{1,opt}\ =\ min\ f_{1}(x)}\\
\bm{f_{2,opt}\ =\ min\ f_{2}(x)}\\
\bm{\cdot}\\
\bm{\cdot}\\
\bm{f_{n,opt}\ =\ min\ f_{n}(x)}
\end{split}
\end{equation}

During optimization, the Pareto technique maintains the components of the solution vectors apart (independent), and the idea of dominance is used to distinguish between dominated and non-dominated solutions. In MOO, the dominant solution and optimal value are often reached when one objective function cannot be increased without decreasing the other objective function. This state is known as Pareto optimality. Pareto optimum solution is the name given to the collection of ideal solutions in MOO. A non-dominated solution, often known as Pareto efficient, is a concept in mathematics. Non-Pareto optimum solutions are those in which one objective function can be increased in such a way so that other's objective function is not affected. This answer is referred to as the dominating solution (worse). Once a Pareto optimal solution is obtained, it can be solved mathematically \cite{r29}. This approach requires taking notice of a number of terms in the Pareto optimum solution. These are the terms: \\
(a) \textbf{Anchor Point}: Through the use of an objective function, anchor points can be found.\\
(b) \textbf{Utopia Point}: The meeting point of the maximum/minimum value of one objective function with the maximum/minimum value of another objective function yields the Utopia point.

\begin{figure}[h]
\centering
\includegraphics[scale=0.5]{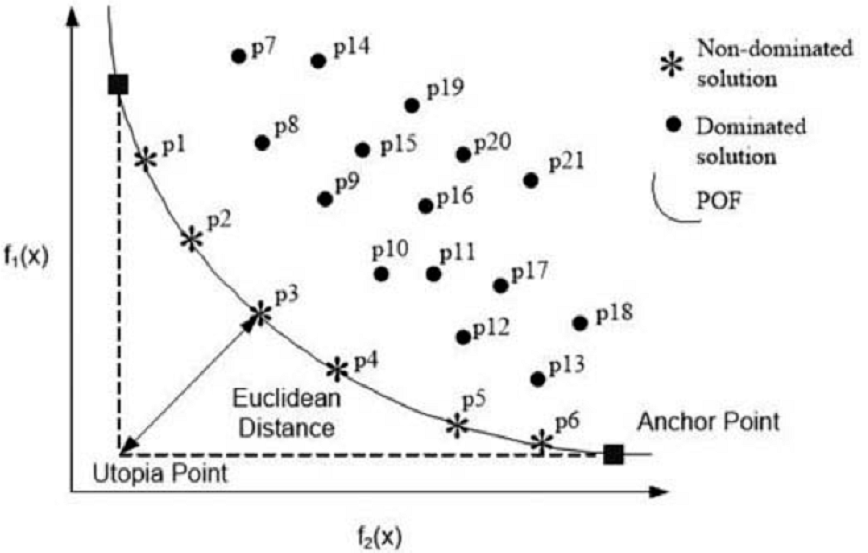}
\caption{ \centering POF of two objective functions (Gunantara, \cite{r70})}\label{fig:fig3}
\end{figure}

\begin{figure}[h]
\centering
\includegraphics[scale=0.5]{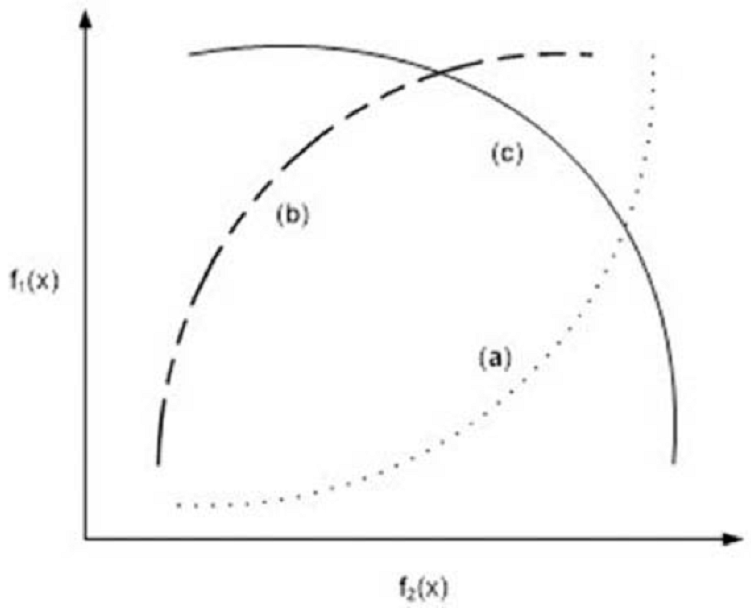}
\caption{\centering Two other objective functions' POF (Gunantara, \cite{r70})}\label{fig:fig4}
\end{figure}

A Pareto optimum front (POF) on a two-dimensional surface can illustrate the optimization with two objective functions and the non-dominated solution \cite{r30}. Consider minimizing $f_1(x)$ and $f_2(x)$ objective functions. Figure.~\ref{fig:fig3} shows the dominated solution (p7, p8,..., p21) and the non-dominated solution (p1, p2, p3, p4, p5, and p6) \cite{r31,r32}. As shown in Figure.~\ref{fig:fig4} under curves (a), (b), and (c), respectively, the POF may also exist in three distinct combinations for minimizing $f_1(x)$ and maximizing $f_2(x)$, maximizing $f_1(x)$ and minimizing $f_2(x)$, and maximizing both $f_1(x)$ and $f_2(x)$. The Pareto optimum solutions' solution sites are depicted in Figure.~\ref{fig:fig3} however. Dominated solutions and non-dominated solutions can be found through the comparison of two solution points for all solution points. For example, the p3 solution is said to be dominant over the p9 solution if the following two conditions are satisfied \cite{r25}:\\
(a) For all objective functions, the p3 solution is not significantly worse than the p9 solution.\\
(b) In terms of one or more than one objective function, the p3 solution triumphs over the p9 solution.\\

\unskip
	
Once the Utopia point has been identified, the shortest Euclidean distance may be used to estimate the POF's ideal value \cite{r33}. The POF form in three-dimensional space can be used to express non-dominated solutions for three objective functions. The non-dominated solutions, however, cannot be seen in the POF if the optimization comprises more than three objective functions \cite{r32}. Use the Continuously Updated approach to look for non-dominated solutions. The quest for non-dominated solutions is ongoing with this strategy. The Continuously Updated technique can \cite{r25}:\\
(a) The path chosen for the start is not dominated by P' = 1. Put counter i to 2.\\
(b) Put j = 1.\\
(c)  In order to find a solution that is more dominant, solutions i and j from P' were compared.\\
(d) If solution i outperforms solution j, then remove the number of -j from P'. If j is smaller than P', multiply it by one and return to step c. Return to step e if the contrary is true. If member number -j from P' dominates member number i, add one to i and return to step b.\\
(e) Replace P' with the answer i or make P' = P'i. Add i with one and return to step b if i$<$N, where N is the number of solutions. If not, the procedure halts and P' is designated as a non-dominated set. The POF is composed of the non-dominated set.\\
The Utopia point is then found out following the completion of the algorithm Continuously Updated. The ideal value is determined using the Utopia point. The shortest Euclidean distance based on equation~\ref{eq4} may be used to obtain the ideal value \cite{r34}. The following equation may be used to get the shortest Euclidean distance between the Utopia point and the POF's points \cite{r34}.\\
After the algorithm Continuously Updated is completed, the Utopia point can be reached. This point can be used to establish the ideal value. Based on equation~\ref{eq4}, the shortest Euclidean distance may be determined \cite{r34} to get the best value. The equation may be applied to get the shortest Euclidean distance between the Utopia point and the POF's points \cite{r34}:
\begin{equation} \label{eq4}
\bm{d_E = min\sqrt{\left(\frac{Q_1 -Q_1^*}{Q_{1norm}} \right)^2 + \left(\frac{Q_2 -Q_2^*}{Q_{2norm}} \right)^2}}
\end{equation}
where, (according to Figure.~\ref{fig:fig3} as an example), ($Q_1^*$,$Q_2^*$) are the coordinates for the Utopia point of the objective function $f_1(x)$ whose minimum value is searched for, and the objective function $f_2(x)$ which needs the minimum value to be determined, ($Q_1$,$Q_2$) are the point coordinates on the POF, and $Q_{1norm}$; $Q_{2norm}$ are normalization point coordinates in the problem areas. $Q_{1norm}$ and $Q_{2norm}$ are determined based on the minimum value of $Q_1$ and the minimum value of $Q_2$ respectively.

where (using Figure.~\ref{fig:fig3} as an example) ($Q_1^*$,$Q_2^*$) are the point coordinates on the POF, ($Q_1$,$Q_2$) are the normalization point coordinates in the problematic regions, and ($Q_{1norm}$; $Q_{2norm}$) are the coordinates for the normalization point in the Utopia points of the objective function $f_1(x)$ whose minimum value is sought after. Based on the minimal values of $Q_1$ and $Q_2$, respectively, $Q_{1norm}$ and $Q_{2norm}$ are calculated.
\subsection{Scalarization method}
The multi-objective function generates a single solution using the scalarization technique. Before the optimization process, the weight is chosen. This approach integrates multiple objective functions into a scalar fitness function using an equation as follows \cite{r64}:
\begin{equation} \label{eq5}
\bm{F(x) = w_1f_1(x) + w_2f_2(x) + \cdots + w_nf_n(x)}
\end{equation}
An objective function weight can determine the fitness function solution and can show the performance priority \cite{r65}. The weight assigned to each objective function in a scalar fitness function determines the priority of each function in the solution. When an objective function is given a higher weight, it gets greater priority when compared to items of lesser weight. There are three ways to determine the scalarization weight: equal weights, ROC weights, and RS weights \cite{r66}. Equal weights can be calculated by using the equation as follows \cite{r67}:

\begin{equation} \label{eq6}
\bm{w_i=\frac{1}{n}}
\end{equation}
where i = 1, 2, 3, , n; given n is the number of objective functions. For ranking different criterion, rank-order centroid (ROC) weights are computed using the equation as in \cite{r68}:
\begin{equation} \label{eq7}
\bm{w_i=\frac{1}{n}\Sigma^{k=i}_n\frac{1}{k}}
\end{equation}
Each criterion is given a proportionate weight using rank-sum (RS) weights. The equation below can be used to calculate RS weights \cite{r68}:
\begin{equation} \label{eq8}
\bm{w_i=\frac{2(n+1-i)}{n(n+1)}}
\end{equation}
In this technique, minimizing function and maximizing function are marked as negative and positive respectively. To make objective functions fair in the scalarization method, it is important to normalize them using the root mean square \cite{r69}. For the three objective functions, a scalarization example is given below:
\begin{equation} \label{eq9}
\bm{F(x)=-\frac{w_1f_1(x)}{\sqrt{E(f_1^2(x))}}+\frac{w_2f_2(x)}{\sqrt{E(f_2^2(x))}}-\frac{w_3f_3(x)}{\sqrt{E(f_3^2(x))}}}
\end{equation}
where F(x) is the fitness function, $f_1(x)$, $f_2(x)$, $f_3(x)$ for objective functions 1, 2, 3 respectively and $w_1$, $w_2$, $w_3$ are the corresponding weights. In MOO, for checking the overall solution, the exhaustive method is used to determine the optimal value. Certain algorithms (such as ant colony optimization, particle swarm optimization (PSO), and meta-heuristic algorithms like GA, etc.) can determine the optimal value, to assist in the optimal solution finding process for a large solution.

\section{Some recent works of deep learning methods with MOO based parameter optimization}
\begin{figure}[htbp]
\centering
\includegraphics[scale=0.40]{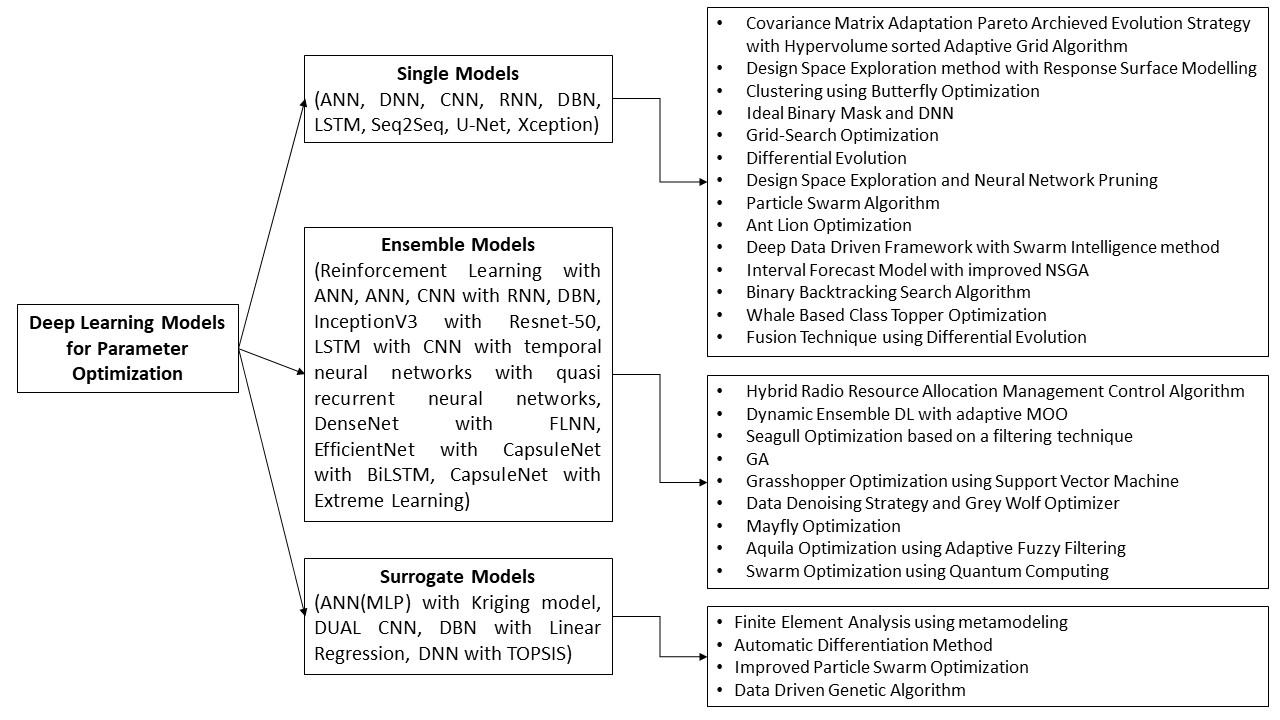}
\caption{\centering Classification of Deep Learning Model based MOO methods \label{fig:fig6}}
\end{figure}

In this section, we provide a summary of the various works of literature as case studies. The deep learning methods and the corresponding frameworks using MOO strategies have been classified in Figure.~\ref{fig:fig6}. The following sections provide an overview of all the methods discussed. We have surveyed about twenty three papers obtained from mainly two major sources, including Google Scholar and Scopus, in the year 2016 to 2022. We have filtered the commonly appearing works in both search spaces based on the following keywords: The primary keywords used were: \textit{deep learning, neural networks, multi-objective optimization, and parameter optimization}. We have classified the models using single-model architectures, followed by ensemble and surrogate models. It contains the key features and the applications of the methods, sorted chronologically.
\subsection{Single Models based Architectures}
In this section, the stand alone DNN architectures using MOO based are provided in Table.~\ref{tab1.1} followed by two of the detailed discussion of few of the methods.
\begin{table}[htbp]
\centering
\caption{Single DNN Model based frameworks}
\resizebox{\columnwidth}{!}{\begin{tabular}{|l|l|l|l|}
\hline
\textbf{Year} &
  \textbf{Algorithm/Model} &
  \textbf{Key Features} &
  Application \\ \hline
2016 &
  \begin{tabular}[c]{@{}l@{}}Design Space Exploration \\ method with Response Surface\\ \cite{r99}\end{tabular} &
  \begin{tabular}[c]{@{}l@{}}- Uses optimization for low-power \\ mobile and embedded areas.\\ - Reduces number of training \\ computations\end{tabular} &
  \begin{tabular}[c]{@{}l@{}}-Used in image\\ processing applications\end{tabular} \\ \hline
2017 &
  \begin{tabular}[c]{@{}l@{}}Multi-objective Evolution \\ of ANN for Medical \\Diagnosis \cite{r44}\end{tabular} &
  \begin{tabular}[c]{@{}l@{}}- Includes Covariance Matrix \\Adaptation's exploration of \\ the problem space \\ - Converges in the direction of \\ true Pareto-optimal front\end{tabular} &
  \begin{tabular}[c]{@{}l@{}}- Used for classification \\ of fetal cardiotocograms\end{tabular} \\ \hline
2017 & \begin{tabular}[c]{@{}l@{}}Deep Learning and Multi-\\ Objective Learning with \\Speech Enhancement \cite{r96}\end{tabular} &  \begin{tabular}[c]{@{}l@{}}
  - incorporates learned \\characteristics, such as the \\ ideal binary mask (IBM) \\ and mel-frequency cepstral \\coefficients (MFCCs), into the \\ original DNN via an auxiliary\\ structure \end{tabular} &
  \begin{tabular}[c]{@{}l@{}}- Used for speech enhancement\\ by reducing speech \\ distortions especially at high \\ SNR frequency units\end{tabular} \\ \hline
2019 &
  \begin{tabular}[c]{@{}l@{}}Traffic flow forecasting using\\ Deep Belief Network and Multi-\\ Objective Particle Swarm \\ algorithm \cite{r86}\end{tabular} &
  \begin{tabular}[c]{@{}l@{}}- Uses feature detectors to\\ extract complex features of\\ traffic flow\end{tabular} &
  \begin{tabular}[c]{@{}l@{}}- Helps in planning better traffic \\ operations\\ - Suggests travelers to select\\ convenient travelling routes\end{tabular} \\ \hline
2019 &
  \begin{tabular}[c]{@{}l@{}}Combustion system based \\deep data-driven models\\ and multi-objective \\ optimization\cite{r97}\end{tabular} &
  - Deep Belief Network used &
  \begin{tabular}[c]{@{}l@{}}- Increase energy efficiency and \\ reduce emissions by optimizing\\ operations\\ - Combustion efficiency and NOx \\emission predictions accurately\end{tabular} \\ \hline
2020 &
  \begin{tabular}[c]{@{}l@{}}Optimization framework \\ in Embedded Systems \cite{r83}\end{tabular} &
  \begin{tabular}[c]{@{}l@{}}- Exploits a constrained design\\ space driven by complex \\architecture\end{tabular} &
  \begin{tabular}[c]{@{}l@{}}- Useful for embedded systems,\\ where memory and energy resources \\ are constrained\end{tabular} \\ \hline
2021 &
  \begin{tabular}[c]{@{}l@{}}MOO Chaotic Butterfly\\ Optimization with DNN\\  \cite{r39}\end{tabular} &
  \begin{tabular}[c]{@{}l@{}}- Energy Efficient Particle Swarm\\ Optimization (PSO) based \\ Clustering is used.\end{tabular} &
  \begin{tabular}[c]{@{}l@{}}- Used to cluster healthcare \\ IOT devices and diagnose\\ disease using collected EHR\end{tabular} \\ \hline
2021 &
  \begin{tabular}[c]{@{}l@{}}Multi-Objective based \\ Differential Optimization for \\ CNN models (MODE-CNN) \cite{r35}\end{tabular} &
  \begin{tabular}[c]{@{}l@{}}- Achieves minimum segmentation\\ loss with shortest test\\ time \end{tabular} &
  \begin{tabular}[c]{@{}l@{}}-Analysis of medical \\ images like segmentation \\ and classification\end{tabular} \\ \hline
2021 &
  \begin{tabular}[c]{@{}l@{}}Multi-objective based \\ wind forecast model \\ using  LSTM model \cite{r82}\end{tabular} &
  \begin{tabular}[c]{@{}l@{}}- Uses a lower and upperbound \\ estimation method to construct \\ Prediction Intervals (PIs)\\ - NSGA is improved by \\ competitive learning mechanism\end{tabular} &
  \begin{tabular}[c]{@{}l@{}}- Constructs prediction intervals \\ for wind power\\ - LSTM based time series \\ feature of wind generation\end{tabular} \\ \hline
2022 &
  \begin{tabular}[c]{@{}l@{}}MOO using Grid Search \\ Optimization with DNN\\ \cite{r40}\end{tabular} &
  \begin{tabular}[c]{@{}l@{}}- Combines Data Missing Care\\ Framework and Grid-Search \\ optimization\\ - Tunes assessments on ANN,\\ CNN and RNN.\end{tabular} &
  \begin{tabular}[c]{@{}l@{}}- Problem of large missing \\ data in medical datasets\\ addressed\\ - Energy management, environment,\\ and medicine\end{tabular} \\ \hline
  2022 &
  \begin{tabular}[c]{@{}l@{}}Wind speed forecasting \\ based Deep learning \\ with Multi-Objective\\ Parameter Optimization\\ \cite{r84}\end{tabular} &
  \begin{tabular}[c]{@{}l@{}}- Effective multi-objective binary\\ backtracking search algorithm \\ (MOBBSA) is employed\\ - Hybrid time series \\decomposition (HTD) used \\ for feature extraction\\ - Advanced Sequence-to-Sequence\\ (Seq2Seq) used for final predictions\end{tabular} &
  \begin{tabular}[c]{@{}l@{}}- Real-world wind speed \\ forecasting\end{tabular} \\ \hline
\end{tabular}}
\label{tab1.1}
\end{table}
\subsubsection{Multi-Objective based Differential Optimization for CNN models:}
The manual tuning of the parameters forms one of the drawbacks for CNN models in achieving high performance. MODE-CNN \cite{r35} focuses on CNN-based optimization of parameters using the multi-objective differential evolution (MODE) algorithm. This is mainly developed for image analysis in the healthcare domain. The accuracy of the model is regulated by the three parameters, which include: patch accuracy, general stride, and neighbor distance.
 Segmentation loss $SL$ is calculated by : 
 \begin{equation} \label{eq10}
\bm{SL= (1- \frac{1}{m}\sum_{i=1}^{m}\frac{1}{a}\sum_{j=1}^{a}(IOU)_{j})}
\end{equation}
where $IoU$ is the ratio of the total number of pixels in the image to the number of pixels where the item estimated using ground truth intersects, $a$ is the total count of objects in the image, $m$ is the total number of images, and $a$ is the total count of objects in the image. \\
Following is the algorithm provided.

\begin{algorithm}\label{alg:alg1}
\renewcommand{\thealgorithm}{}
\caption{of MODE-CNN}
\begin{algorithmic}[1]
\State MODE-CNN initial values are created
\State Appropriate values of general stride (ST), neighbour distance(DIS) and patch accuracy (PAC) with MODE-CNN are created\label{l2}
\State Segmentation error and test time with CNN-based method are found out
\If{$the\ desired\ iteration\ has\ been\ reached$}
    \State \textbf{return} ST, DIS and PAC values
\Else
    \State \textbf{go to} 2
\EndIf
\end{algorithmic}
\end{algorithm}

\textbf{Characteristics:}
\begin{itemize}
\item The score for every individual in a population is calculated using the crowding distance and Pareto front numbers.
\item Roulette wheel selection technique is used for parent selection, which in turn helps to achieve fast convergence.
\item Minimum segmentation loss and minimal test time have been achieved based on the above mentioned parameters using this algorithm.
\item Optimization using this algorithm is achieved in fewer iterations.
\item It was demonstrated to be a robust and competitive algorithm when compared with other popular multi-objective optimization algorithms like NSGA-II, MODE, and others. 
\end{itemize}

\subsubsection{MOO Chaotic Butterfly Optimization with DNN:}
To perform clustering for diagnosing diseases, the MOO based Chaotic Butterfly Optimisation Algorithm with Deep Neural Network (MOCBOA-DNN) technique was developed \cite{r39}. Firstly, a fitness function is used to choose the best set of cluster heads (CHs) and arrange clusters while clustering the IoT medical devices. Secondly, the cloud server receives the gathered medical data for further analysis. The healthcare data is then analysed by the DNN model to determine whether a condition really exists.\\
Following is the Butterfly Optimization Algorithm used in the finding of clusters for the IOT devices. The fragrance $f$ estimate for the can be written as in equation~\ref{eq11} where, $c$ is the sensory modality, $I$ is the intensity of stimuli, and $\beta$ is the power exponent value. These parameters play a significant role where $\beta$ varies between 0 and 1. The value 1 denotes that the neighboring butterfly can sense full fragrance.  In an ideal environment, there is no absorption of fragrance. \\   
\begin{equation}
    \bm{ f= cI^{\beta}}
    \label{eq11}
\end{equation}
The global search where the butterflies move towards the best butterfly can be defined as in Eq. \ref{eq12} where $g*$ is the overall highest value obtained among all solutions in the current iteration, and r is a generated random number. 
\begin{equation}
\bm{ y_i^{t+1} = y_i^t +( r^2 \times g^* - y_i^t ) \times f}
\label{eq12}
\end{equation}
\begin{algorithm}\label{alg:alg5}
\renewcommand{\thealgorithm}{}
\caption{of BOA}
\begin{algorithmic}[1]
\State Generate a population of $n$ butterflies $y_i= (i=1, 2,.., n)$.
\State Initialize Sensor modality $c$, Switch probabilities $p$, and Power exponents $\beta$.
\While{end criteria remains unsatisfied}
\For{all butterflies $b_f$ in the population}
\State Do fragrance estimation as in equation\ref{eq11}
\State Choose best butterfly based on the best solution $(g*)$
\EndFor
\For{all butterflies $b_f$ in the population}
\State  Generate $r$ between 0 and 1
\If{$r < p$}
\State Moves the optimum butterfly near best butterfly $(g^*)$ as in equation \ref{eq12}.
\Else
\State Moves arbitrarily
\EndIf
\State Evaluate for a novel butterfly
\State Upgrade the population when optimum is obtained
\EndFor
\State Upgrade the value of $c$
\State Assign the best overall solution $(g^*)$
\EndWhile
\State Print the optimum solutions $(g*)$
\end{algorithmic}
\end{algorithm}

\textbf{Characteristics:}
\begin{itemize}
 \item Using gathered healthcare data, the MOCBOA-DNN technique may cluster IoT devices for healthcare and diagnose diseases, which can lead to more efficient and effective healthcare management systems.
    \item The best set of cluster heads is chosen using a fitness function for the clustering of IoT medical devices. 
    \item The collected healthcare data present in the cloud storage is analyzed further to check for the presence of any disease. 
    \item When it comes to a diverse variety of evaluation components, the MOCBOA-DNN technique can outperform other current techniques, indicating its effectiveness.
\end{itemize}

\subsubsection{MOO using Grid-Search Optimization with DNN:}
Based on deep-learning optimization models, this framework was created for medical datasets with a high percentage of missing values \cite{r40}. The Data Missing Care (DMC) Framework, which addresses the issue of excessive missing data in medical records, is used to increase the model's resilience. By adjusting multiple hyperparameters, ANN (Artificial Neural Network), CNN (Convolutional Neural Network), and Recurrent Neural Networks (RNN) based deep learning algorithms are tuned, and Grid-Search optimization is utilized to create a better deep predictive training model for patients with COVID-19. \\
 Following is the algorithm for resolving the issue of a large amount of missing data in medical databases. In the pre-processing stage, the data is normalized using the following equation~\ref{eq13} where $p_i$ denotes the original feature value and the minimum and maximum feature values are indicated by $min(p)$ and $max(p)$ respectively:\\
\begin{equation}
\bm{p_i^{new}=\frac{p_i^{old}-min(p)}{max(p)-min(p)}} 
\label{eq13}
\end{equation}

\begin{algorithm}\label{alg:alg6}
\renewcommand{\thealgorithm}{}
\caption{of Multi-Objective Deep learning Framework}
\begin{algorithmic}[1]
\State Data normalized using equation \ref{eq13}
\State Implement ANN, CNN and RNN models
\State Perform hyperparameter tuning using Grid Search Optimization
\State Evaluate prediction results using following:
\State $Accuracy = \frac{(TP+TN)}{(TP+TN+FP+FN)}*100$
\State $Precision = \frac{TP}{(TP+FP)}*100$
\State $Recall = \frac{TP}{(TP+FN)}$
\State $F1\ Score = 2*\frac{Precision*Recall}{Precision+Recall}$
\State Terminate
\end{algorithmic}
\end{algorithm}

\textbf{Characteristics:}
\begin{itemize}
    \item Using the Grid Optimization technique, the deep learning models' hyperparameters were optimized.
    \item The analysis used the following hyperparameters: the number of layers, neurons, activation function, momentum, loss function, learning rate, batch size, and epochs. 
    \item This method can handle the imbalanced datasets commonly found in medical datasets, and the F1 score is used as the ultimate accuracy measuring metric. 
    \item The RNN model performed best in comparison to the CNN and ANN models. 
    \item This technique may be utilized in future studies in the fields of traffic control, electric power networks, financial companies, and other sectors that work with high-dimensional datasets and need rapid data processing.
\end{itemize}

\subsubsection{Multi-Objective Evolution of ANN for Medical Diagnosis:}
A selection mechanism based on the hypervolume indicator is combined with a quick approximate algorithm in the Covariance Matrix Adaptation Pareto Achieved Evolution Strategy with Hypervolume Sorted Adaptive Grid Algorithm (CMA-PAES-HAGA) \cite{r74} to explore the problem space. With the use of variation operators, CMA-PAES-HAGA is able to converge towards the true Pareto-optimal front, \cite{r44} while preserving a diverse population of solutions during the optimization process. \\Following is the Covariance Matrix Adaptation Pareto Archieved Evolution Strategy with Hypervolume sorted Adaptive Grid Algorithm.
\begin{algorithm}\label{alg:alg10}
\renewcommand{\thealgorithm}{}
\caption{of CMA-PAES-HAGA}
\begin{algorithmic}[1]

\State Generation counter and extreme values vector are initialized
\State $t \gets 0$
\State $Z \gets (\varepsilon_{1} = 0, \varepsilon_{2} = 0, \cdots , \varepsilon_{M} = 0)$
\State Parent population is initialized, where X contains the solutions in the search space and Y contains the vectors of objective values
\State Initialise parent population $Y, X$
\While{not met termination criteria}
\For{$j = 1, \cdots, \lambda$}
\State $X_{j}^{'} \gets X_{j}$
\State $X_{j}^{'} \gets X_{j}^{'}+\sigma_j \cdot \mathscr{N}(0, C_j)$
\State Check if solution is within bounds
\If{$X_i^{(L)} \nleq X_{ij}^{'} \nleq X_i^{(U)}$}
\If{$X_{ij}^{'} > X_i^{(U)}$}
\State $X_{ij}^{'}=X_i^{(U)}$
\Else
\State $X_{ij}^{'}=X_i^{(L)}$
\EndIf
\EndIf
\EndFor
\State Evaluate solution 
\State $Y_j^{'} \gets f(X_j^{'})$
\State $Y^{*} = Y \bigcup Y^{'}$
\State Update extreme values
\For{m= 1,..., M}
\If{$Y_{mj}^{*} > \varepsilon_m$}
\State $\varepsilon_m = Y_{mj}^{*}$
\EndIf
\EndFor
\State Selection routine
\State $Y$, $X$ $\gets$ HypervolumeSortedAGA($Y^*$, $Z$)
\State // variation routine
\State CMAParameterUpdate()
\State $t \gets t + 1$
\EndWhile
\end{algorithmic}
\end{algorithm}

\textbf{Characteristics:}
\begin{itemize}
    \item This method addresses class imbalance concerns without requiring the prior integration of knowledge particular to the problem.
    \item In the case of a multi-class classification problem, it takes into account trade-offs between the classification accuracy of each class.
    \item In the case of a multi-class classification problem, it takes into account trade-offs between the classification accuracy of each class.
     \item The minority class recognition is also improved without making assumptions about misclassification costs.
    \item This also presents the decision-maker with a variety of trained ANNs with trade-offs that are evenly dispersed throughout the Pareto front.
\end{itemize}

\subsection{Ensemble Models based Architectures}
In this section, we discuss the different ensemble models using MOO based parameter optimization. Table.\ref{tab2} enlists the ensemble methods using MOO based optimization along with the deep learning algorithms in a sequential or integrated form. We discuss some of the methods in detail with their working frameworks and areas of application, while all the other methods are included in the Table.\ref{tab2}.
\begin{table}[htbp]
\centering
\caption{Ensemble Models based frameworks}
\resizebox{\columnwidth}{!}{\begin{tabular}{|l|l|l|l|}
\hline
\textbf{Year} &
  \textbf{Algorithm/Model} &
  \textbf{Key Features} &
  Application \\ \hline
2019 &
  \begin{tabular}[c]{@{}l@{}}Ensemble Deep Learning \\ with MOO for prediction\\ of treatment outcomes\\ \cite{r41}\end{tabular} &
  \begin{tabular}[c]{@{}l@{}}- Trains with deep perceptron \\ models to handle issues\\ of EHR data\\  - Ensemble strategy with \\ adaptive multi-objective \\optimization and evidential \\ reasoning (ER) fusion used\end{tabular} &
  \begin{tabular}[c]{@{}l@{}}- Prediction of treatment risks\\ after radiotherapy for\\ lung cancer\\ - Identifies features like tumor \\ size, regional dose, T staging, \\ N staging through feature \\ importance analysis\end{tabular} \\ \hline
2021 &
  \begin{tabular}[c]{@{}l@{}}Multi-Objective Ensemble \\ Deep Learning for prognosis \\ of rotating machinery \cite{r81}\end{tabular} &
  \begin{tabular}[c]{@{}l@{}}- Monitoring data collected by \\ prognostics and health  management \\ (PHM)\end{tabular} &
  \begin{tabular}[c]{@{}l@{}}- Monitoring data used\\ to calculate Remaining Useful \\ Life  (RUL) of mechanical components \\ - Applied in PHM systems using \\ smart sensing techniques and\\ the Internet of Things (IoT)\end{tabular} \\ \hline
2021 &
  \begin{tabular}[c]{@{}l@{}} Character Recognition based\\ Aquila Optimizer and DNN  \\ \cite{r93}\end{tabular} &
  \begin{tabular}[c]{@{}l@{}}- Fuzzy filtering technique for\\ image pre-processing\\ - Fusion of DNN models \\EfficientNet and CapsuleNet \\for feature extraction\\ - Bi-directional long\\ short-term memory (BiLSTM)\\ model with Aquila optimizer\\ is used\end{tabular} &
  \begin{tabular}[c]{@{}l@{}}- Character recognition, where \\ handwritten and printed  text\\  coexist in the same document\\ - Digitization of historical\\ documents containing Telugu\\ characters.\\ - Accessibility for visually\\  impaired\\ - License plate recognition,\\ signboard recognition, and\\ text translation\end{tabular} \\ \hline 
2022 &
  \begin{tabular}[c]{@{}l@{}}Breast Lesion Assesment\\ using combined DNNs and\\ MOO based Seagull\\ Optimization\cite{r42}\end{tabular} &
  \begin{tabular}[c]{@{}l@{}}- Convolutional and Recursive\\ neural networks are \\combined\end{tabular} &
  \begin{tabular}[c]{@{}l@{}}- Identification and categorization \\ of breast lesion\\ - Promising tool for disease\\ prediction, diagnosis, screening\\ and treatment\end{tabular} \\ \hline
2022 &
  \begin{tabular}[c]{@{}l@{}}Multi Objective Grasshopper\\ Optimization Algorithm\\ (MOGOA) \cite{r37}\end{tabular} &
  \begin{tabular}[c]{@{}l@{}}-Two pre-trained CNN\\ architecture such as InceptionV3 \\ and ResNet50 were used.\end{tabular} &
  \begin{tabular}[c]{@{}l@{}}- Classifying the Non-COVID-19,\\ COVID-19, and pneumonia patients\\ using Chest X-ray images\end{tabular} \\ \hline
2022 &
  \begin{tabular}[c]{@{}l@{}}Short-term wind speed \\ forecasting based on Deep\\ Learning and Multi-objective\\ parameter optimization\\ \cite{r85}\end{tabular} &
  \begin{tabular}[c]{@{}l@{}}- Integrates several single-model\\ forecasting results through a weight\\ optimization operator\\ - Able to provide both point\\ prediction and uncertainty\\ forecasting\end{tabular} &
  \begin{tabular}[c]{@{}l@{}}- Provides accurate and real-time\\ wind power information.\\ - Useful for planning and\\ managing wind power projects\\ - May be applicable in aviation\\ or marine transportation.\end{tabular} \\ \hline
2022 &
  \begin{tabular}[c]{@{}l@{}}Multi-Objective Mayfly \\ Optimization with DenseNet\\ \cite{r92}\end{tabular} &
  \begin{tabular}[c]{@{}l@{}}- Involves DenseNet-169 as a feature\\ extractor.\\ - Uses functional link neural\\ network (FLNN) as a classification\\ model.\end{tabular} &
  \begin{tabular}[c]{@{}l@{}}- Assistive technologies for visually \\ impaired people\\ - Automated registry for\\ business documents\\ - Real-time handwritten recognition \\ in smartphone environment\end{tabular} \\ \hline
2022 &
  \begin{tabular}[c]{@{}l@{}}Multi-Objective Quantum\\ Swarm Optimization with \\DNN \cite{r36}\end{tabular} &
  \begin{tabular}[c]{@{}l@{}}- Involves optimized region\\  growing based segmentation.\\ - Involves capsule network \\ (CapsNet) based feature extraction.\\ - Involves extreme learning \\ machine (ELM) based\\ classification.\end{tabular} &
  \begin{tabular}[c]{@{}l@{}}- Diagnose dystrophinopathies\\ using muscle Magnetic\\ Resonance Imaging (MRI)images\end{tabular} \\ \hline
\end{tabular}}
\label{tab2}
\end{table}
\subsubsection{Multi-Objective Mayfly Optimization with DenseNet:}
\begin{figure}[h]
\centering
\includegraphics[scale=0.35]{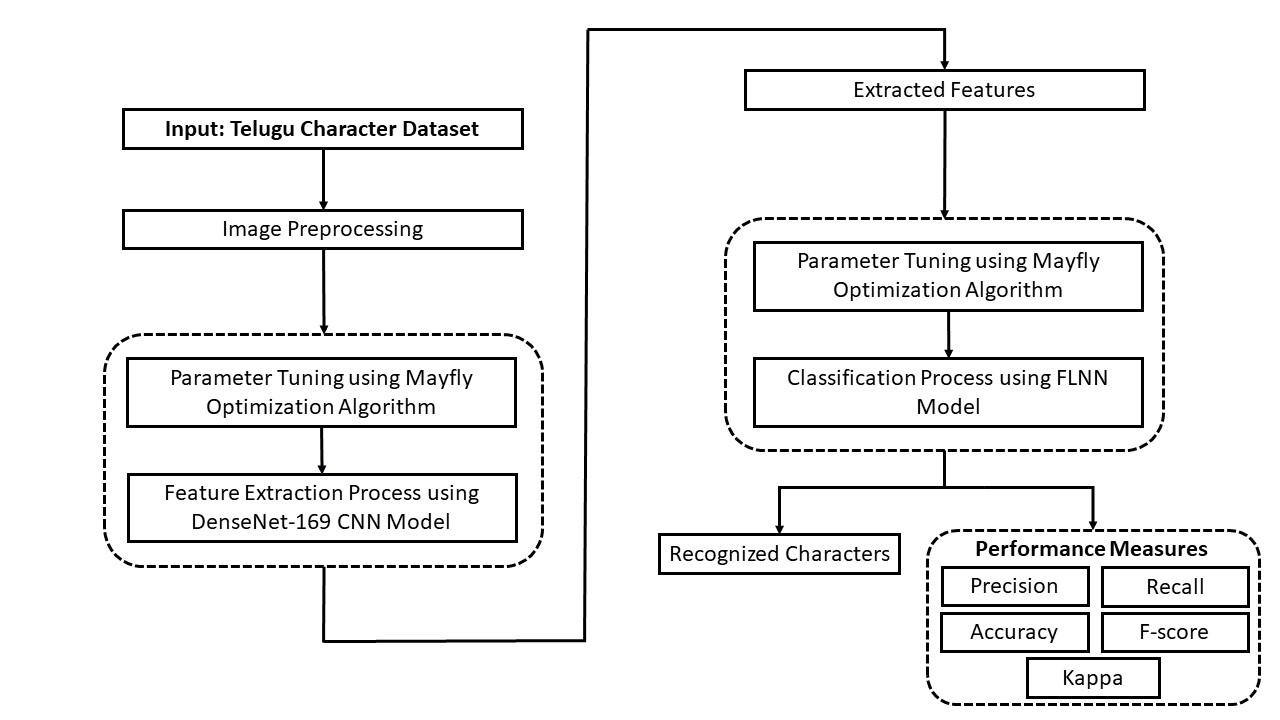}
\caption{\centering Overall Architecture of Proposed Model\cite{r92}}\label{fig:Case22}
\end{figure}

As a feature extractor, a DenseNet-169 model is utilized to generate a set of beneficial feature vectors in the Multi-Objective Mayfly Optimisation with Deep Learning (MOMFO-DL) technique, and a Functional Link Neural Network (FLNN) is used for classification to identify and categorize the handwritten characters \cite{r92}. The DenseNet model and FLNN model parameters are optimized using the MOMFO method. The overall architecture is shown in Figure.~\ref{fig:Case22}. Recognition of handwritten characters has gained prominent attention in recent times due to its wide application in various technologies like providing assistance to visually impaired people, automating the registry of business documents, and others. This particular work focuses on Telugu handwritten character recognition. Following is the algorithm of Mayfly Optimization.

\begin{algorithm}\label{alg:alg22}
\renewcommand{\thealgorithm}{}
\caption{of Mayfly Optimization}
\begin{algorithmic}[1]
\State Male and Female Mayfly Population is initialized
\State Upgradation of the Velocities and Solution
\State Upgradation for male mayflies
\If{$f(b_i)>f(b_{h_i})$}
    \State $s_i(t+1)=g \cdot s_i(t)+a_1e^{-\beta r_p^2}[b_{h_i}-b_i(t)]+a_2e^{-\beta r_q^2},[b_g-b_i(t)]$
\Else
    \State $s_i(t+1)=g \cdot s_i(t)+d \cdot r_1$
\EndIf
\State where,
\State $f(b_i)$ indicates present fitness value
\State $f(b_{h_i})$ indicates optimal fitness value in trajectories
\State $g$ indicates linearly declination of small one from the maximal values
\State $a_1$, $a_2$ and $\beta$ are constants used to balance the values
\State $r_p$ and $r_g$ denotes parameters used for calculation of Cartesian distance among the individuals and their finest global and previous positions in swarm
\State d is an arbitrary dance coefficients
\State $r_1$ indicates a random quantity in uniform distributions from [1, 1]
\State Upgradation for female mayflies
\If{$f(c_i)<f(b_i)$}
    \State $s_i(t+1)=g \cdot s_i(t)+a_3e^{-\beta r_{mf}^2}[b_i(t)-c_i(t)]$
\Else
    \State $s_i(t)=g \cdot s_i(t)+fl \cdot r_2$
\EndIf
\State where,
\State $a_3$ indicates additional constant
\State $r_m$ indicates Cartesian distance among them
\State $fl$ is additional arbitrary dance
\State $r_2$ indicates a random quantity in uniform distributions from [1, 1]
\State Mayflies are ranked
\State Mayflies are mated and Offspring is evaluated
\State Worst Solutions are replaced with Best New One
\If{Met is terminated}
    \State End
\Else
    \State \textbf{go to} 2
\EndIf
\end{algorithmic}
\end{algorithm}

\textbf{Characteristics:}
\begin{itemize}
    \item The new MOMFO-DL model attempts to enhance the understanding of handwritten Telugu characters.
    \item Pre-processing, feature extraction, classification, and parameter optimization are only a few of the various steps of operations that the model entails.
    \item Most of this approach's inspiration comes from the behaviour of the mayflies. The functioning of the female and male mayflies is implemented for the optimizations, much like the Swarm optimization connected to the swarm individuals. As a result, the classification process has been improved by the application of the May Fly Optimisation method.
    \item The results of the experiment showed better recognition performance with high accuracy, and these findings might be applied in the future to feature selection and segmentation strategy design. They may also be useful in assisting smartphone users.
\end{itemize}
\subsubsection{Multi-Objective Quantum Swarm Optimization with DNN:}
\begin{figure}[h]
\centering
\includegraphics[scale=0.35]{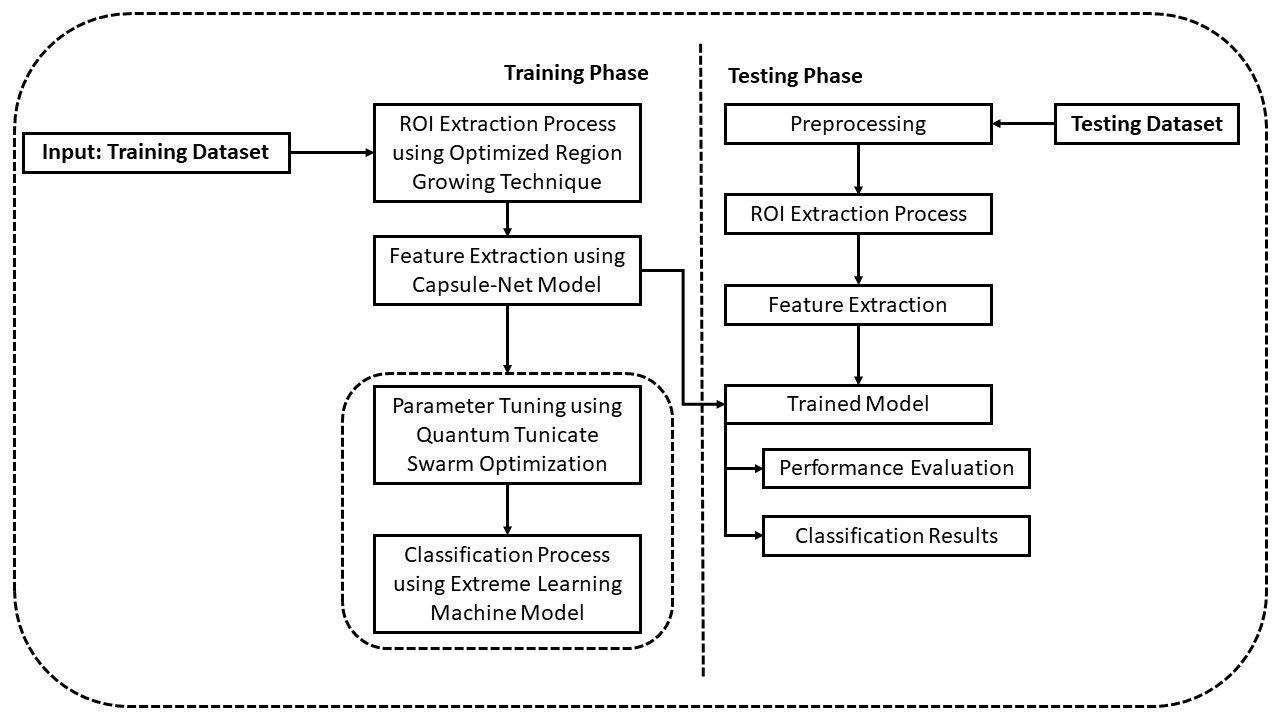}
\caption{\centering Overall framework of MOQTSO-DL method\cite{r36}}\label{fig:Case2}
\end{figure}

The MOQTSO-DL (Multi-Objective Quantum Tunicate Swarm Optimisation with Deep Learning) model by \cite{r36} includes four steps: segmentation based on optimized region growing, feature extraction based on CapsNet, classification based on Extreme Learning Machine, and parameter optimization based on MOQTSO. This algorithm is mostly used to diagnose dystrophinopathies. It forms one of the most commonly inherited muscular diseases across the globe. For this study, muscle MRI images are utilized, and the region of interest (RoI) detection method is predominantly carried out using an optimized region-growing approach. The feature vectors are extracted utilizing the CapsNet model. The ELM classifier then uses the retrieved feature vectors as input to derive the appropriate class labels. The MOQTSO technique is primarily used to choose the first RoI detection seed sites and to fine-tune the ELM model's parameters. Figure.~\ref{fig:Case2} refers to the overall process of the MOQTSO-DL model.

\textbf{Characteristics:}
\begin{itemize}
\item The entire method starts with Region of Interest (RoI) detection, followed by feature extraction using the CapsNet model. \item This method helps to emphasize the relations among the regions of the image and is hence very useful for feature extraction.
\item The feature vectors extracted are then fed to the ELM model for classification of the muscle MRI images. 
\item Finally, the Tunicate Swarm Optimization method (TSO) method is used for parameter selection. The Tunicate Swarm Optimization algorithm is described below. 
\item MOQTSO algorithm reduces time complexity compared to traditional random parameter searching processes.
\item It results in a maximal performance with high accuracy on the classification of Duchenne Muscular dystrophy (DMD) and Becker Muscular Dystrophy (BMD)  versus others.
\item This method can be extended for designing lightweight deep learning models and hyper-parameter optimizers like Adam and others to reduce space-time complexity.
\end{itemize}

\begin{algorithm}\label{alg:alg2}
\renewcommand{\thealgorithm}{}
\caption{ of MOQTSO}
\begin{algorithmic}[1]

\State Initialization of population of tunicates $\vv{P_p}$
\State Original value for parameter and maximum iterations is set
\State Fitness value of every exploration agent are calculated
\State After fitness evaluation, find the optimal entity in the search space 
 Calculate the State Fitness cost of upgraded search agent

\State  Every exploration agent's location is upgraded
\State Novel upgraded agent is returned to their borders
\State Fitness cost of upgraded search agent is calculated 
\If{solution is more optimal than previous}
\State $\vv{P_p}$ is upgraded and the optimal solution is stored in $X_{best}$
\EndIf
\If{end condition is not encountered}
\State iterate steps 5 to 8
\EndIf
\State The optimum solution, which is attained until now, $(X_{best})$ is stated

\end{algorithmic}
\end{algorithm}

\subsection{Surrogate Models based Architectures}
In this section, we mention surrogate model based frameworks using parameter optimization. Table.\ref{tab3} contains the various methods used in the different applications of machinery. 
\begin{table}[htbp]
\centering
\caption{\centering Surrogate Models based frameworks}
\resizebox{\columnwidth}{!}{\begin{tabular}{|l|l|l|l|}
\hline
\textbf{Year} &
  \textbf{Algorithm/Model} &
  \textbf{Key Features} &
  Application \\ \hline
2019 &
  \begin{tabular}[c]{@{}l@{}}DL based Aerodynamic Design\\ optimization \cite{r95}\end{tabular} &
  \begin{tabular}[c]{@{}l@{}}- Multi-fidelity based\\ optimization is used \\ - Deep Belief Network is\\ employed as the low-fidelity model\\ - K-step contrastive divergence\\ algorithm is used for training\end{tabular} &
  \begin{tabular}[c]{@{}l@{}}- Aerodynamic design optimization\\ under 
 uncertainty of Mach number\end{tabular} \\ \hline
2020 &
  \begin{tabular}[c]{@{}l@{}}Design of  Magnet Synchronous\\ Motor for Electric Vehicle based\\ on MOO and DL \cite{r89}\end{tabular} &
  \begin{tabular}[c]{@{}l@{}}- Uses Multi-layer perceptron\\ (MLP) for shape optimization\\ of motors\\ - Utilising finite element\\ analysis to design\\ experiments \end{tabular} &
  \begin{tabular}[c]{@{}l@{}}- Shape optimization of permanent \\ magnet synchronous motor\\  (PMSM)\\ - Maximize PMSM performance\end{tabular} \\ \hline 
2021 &
  \begin{tabular}[c]{@{}l@{}}Aerodynamic prediction using\\ dual CNN and optimization of\\ turbine rotor \cite{r88}\end{tabular} &
  \begin{tabular}[c]{@{}l@{}}- Dual-CNN for the aero-engine\\ turbines\\ - Gradient-based MOO with\\  efficiency and torque as the\\  objective functions\end{tabular} &
  \begin{tabular}[c]{@{}l@{}}-Field reconstruction and \\ performance prediction for a\\ compact turbine rotor in aerospace.\\ - Real-time adjustment direction\\  for operation and maintenance\\  of the rotor .\\ - Predicts pressure and temperature \\ distribution in large\\ gradient areas at leading and\\  trailing edge of blade\end{tabular} \\ \hline
2022 &
  \begin{tabular}[c]{@{}l@{}}MOO of machining process\\ parameters with DNN\\ \cite{r91}\end{tabular} &
  \begin{tabular}[c]{@{}l@{}}- Uses DL for data-driven prediction\\  of optimized objectives.\\ - Uses Technique for Order\\ Preference by Similarity to Ideal\\  Solution (TOPSIS)\end{tabular} &
  \begin{tabular}[c]{@{}l@{}}- Making improved product quality,\\ efficiency, and reduced\\ environmental impact.\\ - Guide operator for machining\\ process parameters selection\end{tabular} \\ \hline
\end{tabular}}
\label{tab3}
\end{table}
\subsubsection{MOO of machining process parameters with DNN:}
\begin{figure}[h]
\centering
\includegraphics[scale=0.5]{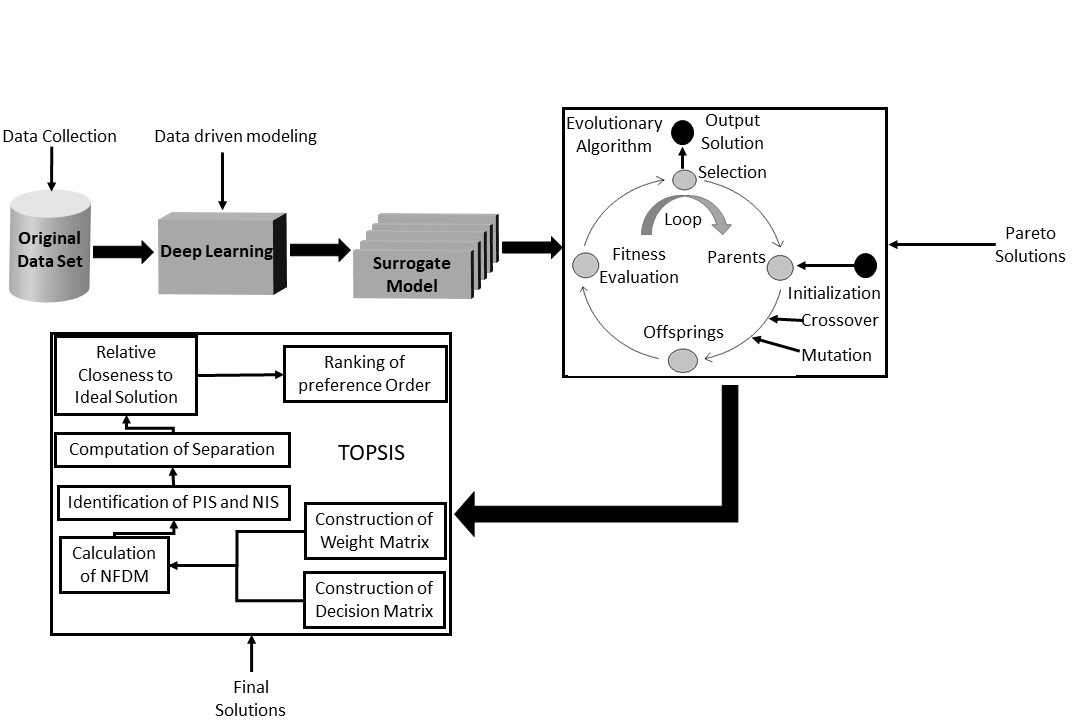}
\caption{\centering Diagram of the overall framework\cite{r91}}\label{fig:Case21}
\end{figure}

Production efficiency and the environmental impact of the machining process are hugely impacted by the machining process parameters \cite{r91}. Existing research employs costly physical models and computationally intensive numerical simulations that are ineffective and inaccurate during the real exploitation stage. For the multi-objective optimization of machining process parameters, a deep-learning based, genetic algorithm as well as TOPSIS-the Technique for Order Preference by Similarity to the Ideal Solution is employed. The overall framework is shown in Figure.~\ref{fig:Case21}. DNN first creates the data-driven prediction function for various optimized objectives. The created prediction function is subsequently transformed into a surrogate model and combined with the genetic algorithm in order to generate the Pareto set. The optimal processing parameter is then automatically found from the generated Pareto set using TOPSIS. The algorithm is provided below.\\

\begin{algorithm}\label{alg:alg21}
\renewcommand{\thealgorithm}{}
\caption{of TOPSIS}
\begin{algorithmic}[1]
\State Normalized decision matrix (NDM) is constructed using $r_{ij}=\frac{x_{ij}}{\sqrt{\sum_{i=1}^{m}x_{ij}^{2}}}\forall{j}$ where $r_{ij}$ indicates the elements of NDM
\State Weighted NDM is constructed using $c_{ij}=r_{ij}*\omega_j\forall{i,j}$ where $\omega_j$ indicates the assigned weight to attribute $j$
\State Idea($X^+$) and negative-idea($X^-$) solutions are determined using \\
    $\biggl\{\begin{pmatrix}max \ c_{ij}\\j\end{pmatrix}|\ i \ \epsilon \ Y,\begin{pmatrix}min \ c_{ij}\\j\end{pmatrix}|\ i \ \epsilon \ Y^{'}$$;\forall{j}\biggl\}=\{c_1^{+},c_2^{+},....\}$ \\
    $\biggl\{\begin{pmatrix}min \ c_{ij}\\j\end{pmatrix}|\ i \ \epsilon \ Y,\begin{pmatrix}max \ c_{ij}\\j\end{pmatrix}|\ i \ \epsilon \ Y^{'};\forall{j}\biggl\}=\{c_1^{-},c_2^{-},....\}$ \\
    where $Y$ and $Y^{'}$ are associated with benefit and cost attributes respectively
\State Separation measure is calculated using 
    $S_i^+=\sqrt{\sum_{i=1}^{n}(c_{ij}-c^+)^2}\forall{j}$, 
    $S_i^-=\sqrt{\sum_{i=1}^{n}(c_{ij}-c^-)^2}\forall{j}$
\State Relative closeness to the ideal solutions is calculated using $C_j^+=\frac{S_j^+}{S_j^++S_j^-}$, alternative rankings based on $C_j^+$ values
\end{algorithmic}
\end{algorithm}

\textbf{Characteristics:}
\begin{itemize}
\item NSGA-III is used as the MOO based genetic algorithm is used for generating Pareto optimal solutions \cite{r108}\cite{r109}.
\item The machining based optimization objectives involve conflicting goals, including roughness of the surface, rate of production, maximal cutting force, and energy consumption 
\item Primary advantages include cost-efficient generic objective functions.
\item An end-to-end structure that minimizes human interference and provides fast optimization speed that takes only a few minutes. 
\item The future scope would include performance degradation, which would be inventible if using such advanced optimization strategies. 
\end{itemize}
\section{Applications}
Multi objective optimization is incorporated with deep learning models in various fields. They are summarized below:

\subsection{Text Recognition:}
Aquila Optimization using Adaptive Fuzzy Filtering \cite{r93} and Multi-Objective Mayfly Optimization with Deep Learning (MOMFO-DL) \cite{r92} techniques aims to detect and recognize handwritten Telugu characters. Character recognition is performed in presence of both handwritten and printed text coexisting in the same document. It is also potentially helpful for creating assisted technologies for visually impaired people. The scope of such works may also be extended to automated registry of documents, signboard and license plate recognition.

 \subsection{Traffic Management:}
 In the case of traffic management, the main purpose is to detect traffic incidents automatically and actively control the traffic based on predictions. A MOO framework using Particle Swarm Algorithm achieves traffic flow forecasting for the next day to actively control and adjust traffic flow, as well as develop plans for managing traffic \cite{r86}.

 \subsection{Wind Speed Forecasting:}
 Interval Forecast Model with improved NSGA plays a crucial part in operating and dispatching contemporary power systems by effectively quantifying uncertainties in wind power forecasting \cite{r82}. A MOO framework using Binary Backtracking Search Algorithm is used in real-world wind speed forecasting \cite{r84}. A MOO framework using Data Denoising Strategy and Grey Wolf Optimizer assists in accurately forecasting short-term wind speeds, which mitigates the effects of wind speed fluctuations \cite{r85}. This aids the decision makers in planning and operators of the power grid systems dispatching power systems in a timely manner, as well as reducing the risk of failure in wind power systems and improving overall power quality.
 
\subsection{Mobile and Embedded Systems:}
The time to market and NRE cost for mobile and embedded systems paves the utility for deep learning based technologies. Design Space Exploration method with Response Surface Modelling automates the artificial neural networks design process, reducing time and costs \cite{r99}. A MOO framework with Neural Network Pruning for exploring Design Space is used to optimize DNNs in embedded systems, which are mostly restricted by constraints in memory and energy resources \cite{r83}.

 \subsection{Machinery:}
By analyzing monitored data collected by PHM systems, a MOO framework employing GA is used to predict the Remaining Useful Life (RUL) of mechanical components \cite{r81}.  Surrogate based modeling systems based frameworks are used mainly for aerdynamics and motor systems. A MOO framework using improved Particle Swarm Optimization is utilized for optimization of aircraft airfoils and wings when the Mach number is uncertain \cite{r95}. Again, Finite Element Analysis employing the metamodeling technique optimizes the design of the permanent magnet based synchronous motors for electric vehicles \cite{r89}. Another MOO framework using Automatic Differentiation Method optimizes turbine rotor by aerodynamic performances prediction for aerospace engineering application \cite{r88}. Data Driven Genetic Algorithm is used in mechanical manufacturing systems to select the optimal machining process parameters, which can effectively enhance the production efficiency of the process and reduce its environmental impact \cite{r91}.

\subsection{Environment Control:}
Several environmental hazards can be overcome with the use of parameter optimization based deep learning methodogies. Deep Data Driven Framework with Swarm Intelligence method improves the combustion efficiency in coal-fired thermal power plants \cite{r97}. It also reduces the NOx emission, thereby reducing its environmental pollution.

\subsection{Healthcare:}
In healthcare, different types of analysis are involved based on the multimodal data available. Covariance Matrix Adaptation based Evolution Strategy with Adaptive Grid Algorithm is used to classify fetal cardiotocogorams by optimizing overall classification accuracy and individual target class accuracy in\cite{r44}. This method has been used for the development of a support based system for decision management for the computerized analysis of fetal cardiotocograms. 
The valuable patient-specific data found in electronic health records (EHRs) can be used to enhance outcome prediction. Multi-objective ensemble deep learning (MoEDL) is devised to predict the probability of significant treatment failure following radiation in lung cancer patients \cite{r41}. Specifically, deep perceptron networks are utilized as base learners throughout the training phase to address a variety of EHR (electronic health record) data-related issues. For the devices in healthcare, machine learning algorithms have widespread applications. Clustering using Butterfly Optimization technique has the objective of categorizing healthcare IoT devices into clusters while analyzing the collected healthcare data to diagnose diseases \cite{r39}. Missing data forms one of the commonly faced issues in the medical area of research. A MOO framework using Grid-Search Optimization is utilized to address the shortcomings of high levels of missing data in medical datasets \cite{r40}. These methodologies have also been used in speech enhancement technologies. A MOO framework using an ideal Binary Mask and DNN is used to enhance speech signals \cite{r96}.
\subsection{Image processing:}
A MOO framework using Differential Evolution is used in segmentation and classification of medical images \cite{r35}. Swarm Optimization using Quantum Computing is used to diagnose dystrophinopathy disease by employing Magnetic Resonance Imaging (MRI) tool \cite{r36}. Grasshopper Optimization using Support Vector Machine automatically uses chest X-ray images to categorize patients under Non-COVID-19, COVID-19, and pneumonia categories \cite{r37}. CT scans and MRI images are also used for image analysis. Based on a combined CNN and RNN framework with optimized parameters, the multi-objective seagull optimization algorithm (BLIC-CRNN-MOSOA) can be used in mammography screening to identify and classify breast lesions. It provides three categories of output: (i) normal, (ii) benign, and (iii) malignant tumors \cite{r42}. The MOSOA method can be used to help the CRNN in finding the best parameters and also for fine-tuning them.

\section{Discussion and Conclusion}
The performance advantages that can be obtained by applying multi-objective based parameter optimization can be more fully utilized while deep learning is an active research topic.  Deep learning currently has two significant limitations. Problems with training data come first, followed by the explainability of these black-box models. The difficulty in mapping the input to the output results in naming such DNNs as black boxes. Other challenges include noisy data, the problem of missing or incomplete data, and others. \\
Presently, there are lots of ongoing works for addressing these drawbacks. Many interpretation and attribution-based explanation strategies are being proposed in the recent literature. In this survey, it is discussed how useful multi-objective optimization is for deep learning parameter optimization. Taking into account the benefits of multi-objective optimizations, such as acquiring improved results with low error, using a smaller data set for the training phase, optimal architectures, and obtaining multiple final solutions, where it is not dependent on an external constraint for only one solution to be deemed the best solution of the Pareto front. It has been seen that using such MOO based parameter optimization strategies in deep learning models can improve their performance, especially for cases with incomplete, missing, and noisy data.\\
 To achieve performance gains,  parameter optimization must be implemented correctly. In comparison to the single objective optimization methods, MOO methods overcome the problems by having lower variance because they intelligently search the parameter space. The widespread applications, particularly in the domains of healthcare and machinery, provide very positive future utilities of such methods, especially with the embedded and surrogate models.
Therefore, future research can focus on enhancing the performance of deep learning models via MOO-based parameter optimization.

\bibliographystyle{cs-agh}
\bibliography{references}

\end{document}